\documentclass[10pt,twocolumn,letterpaper]{article}

\usepackage[pagenumbers]{cvpr} %

\usepackage{cuted} %

\usepackage{currfile} %

\usepackage{caption} %

\usepackage{multirow}
\usepackage{adjustbox}
\usepackage{placeins}
\usepackage{float}
\usepackage{pifont}
\usepackage{tikz}
\usetikzlibrary{positioning,arrows.meta,shapes.geometric,fit,backgrounds,calc}
\newcommand{\cmark}{{\color{ForestGreen}\ding{51}}}
\newcommand{\xmark}{{\color{BrickRed}\ding{55}}}
\newcommand{\pmark}{{\color{orange}$\circ$}}
\newcommand{\objtag}[2]{{\setlength{\fboxsep}{1pt}\colorbox{#1!18}{\textsf{\scriptsize\strut #2}}}}
\newcommand{\tLLM}{\objtag{gray}{LLM}}
\newcommand{\tMLLM}{\objtag{violet}{MLLM}}
\newcommand{\tGame}{\objtag{orange}{Game agent}}
\newcommand{\tCUA}{\objtag{blue}{CUA}}
\newcommand{\tVLA}{\objtag{brown}{Policy}}
\newcommand{\tRUA}{\objtag{green}{RUA}}
\newcommand{\mT}{\textsf{T}}
\newcommand{\mI}{\textsf{I}}
\newcommand{\mV}{\textsf{V}}
\newcommand{\ours}{\textsc{ICLWorld}\xspace}      %
\newcommand{\oursplain}{ICLWorld\xspace}          %
\newcommand{\harness}{\textsc{UIverse}\xspace}   %
\newcommand{\harnessplain}{UIverse\xspace}
\definecolor{tvgray}{gray}{0.88}
\DeclareRobustCommand{\tv}[2]{{\setlength{\fboxsep}{1.2pt}\colorbox{tvgray}{\strut#1}}\,/\,#2}
\newcommand{\app}[1]{Appendix~#1}
\newcommand{\vR}{\textsc{Replay}\xspace}
\newcommand{\vI}{\textsc{Resample}\xspace}
\newcommand{\vE}{\textsc{Relocate}\xspace}

\usepackage{microtype}

\usepackage{float}
\usepackage{array}

\definecolor{cvprblue}{rgb}{0.21,0.49,0.74}
\usepackage[pagebackref,breaklinks,colorlinks,allcolors=cvprblue]{hyperref}

\def\confName{CVPR}
\def\confYear{2027}

\title{ICLWorld: Benchmarking In-Context Learning of Agents from Digital to Physical Worlds}

\author{
Qingle Liu \quad
Jieming Li \quad
Pei-lin Li \quad
Shenghan Zheng \quad
Ruize Yang \\
Sunqi Fan \quad
Deyao Hong \quad
Huan-ang Gao \quad
Meng-Hao Guo \quad
Shuojin Yang$^{\dagger}$ \\[4pt]
Tsinghua University \\[4pt]
{\small Project page: \href{https://iclworld.github.io}{\texttt{iclworld.github.io}} \quad Code: \href{https://github.com/ICLWorld/ICLWorld}{\texttt{github.com/ICLWorld/ICLWorld}}} \\
{\small Dataset: \href{https://huggingface.co/datasets/Aoraku/ICLWorld}{\texttt{huggingface.co/datasets/Aoraku/ICLWorld}}} \\[2pt]
{\small $^{\dagger}$Corresponding author}
}

\begin{document}
\maketitle
\begin{strip}
\centering
\includegraphics[width=\linewidth]{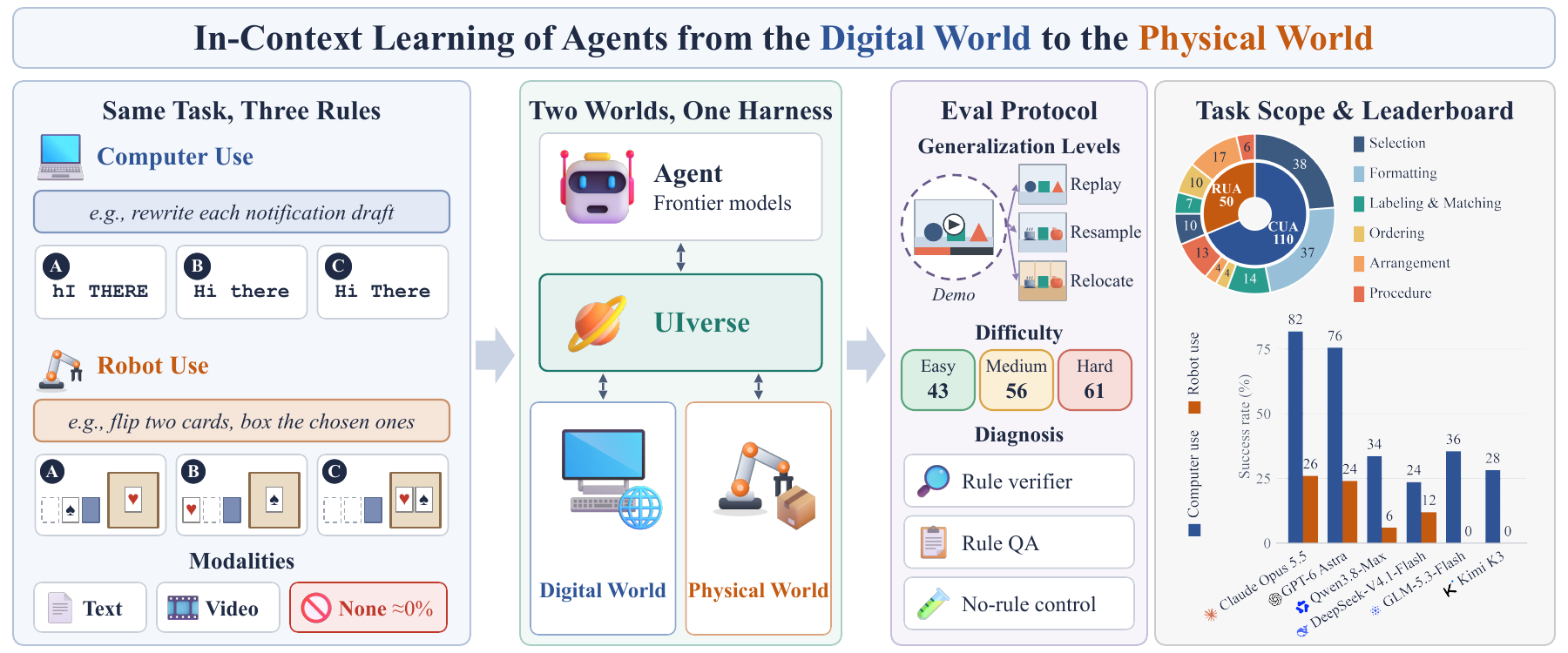}
\captionsetup{hypcap=false}
\captionof{figure}{\textbf{Overview of \oursplain.} Each task comes in three rule versions that require different behavior for the same goal, and the rule is given as text, as a video, or not at all. The same agent acts in the digital and the physical world through one harness, \harnessplain, and is evaluated with rule-specific verifiers, rule questions, and a no-rule control across three difficulty levels and three generalization levels. Right: the 160 tasks by domain and rule type, and the success rates of six frontier agents with text rules.}
\captionsetup{hypcap=true}
\label{fig:teaser}%
\end{strip}

\begin{abstract}
In-context learning (ICL) lets agents acquire new rules from context and adapt to user-specific requirements at inference time.
The same frontier agents now operate both computers and robots, yet their ICL remains insufficiently assessed: existing benchmarks are largely offline, rely on text or static images, and risk training-data contamination, while robot evaluations are limited to method-level ablations.
We present \textbf{ICLWorld}, an online benchmark that evaluates the ICL of agents from digital to physical worlds, covering computer use and robot use, together with \textbf{UIverse}, a unified harness that turns every environment into a user interface.
We build 30 environments, record 630 demonstration videos (10+ hours), and design 160 tasks with customized, often counter-intuitive rules that cannot be satisfied without learning from the context.
Each task comes with three rule versions, given as text or video, and we test generalization at three levels, from replaying the demonstration to new instances and unseen environments.
Across six frontier models, the strongest agents, Claude Opus 5.5 and GPT-6 Astra, succeed on 82\% and 76\% of computer-use runs but only 27\% and 26\% of robot-use runs, and a no-rule control reduces success to at most 8\%.
GPT-6 Astra succeeds under all three rule versions on only 38\% of computer-use tasks, half its rate under a single rule, and answers 98\% of robot rule questions correctly yet completes only 20\% of these tasks.
Frontier agents learn rules comparably well from video and from text and transfer them across environments, whereas weaker agents rely on text and lose most of their video-based success once the demonstrated situation is resampled.
These findings reveal substantial gaps between the digital and the physical world and between identifying a rule and acting on it.
\end{abstract}

\section{Introduction}
\label{sec:intro}

In-context learning (ICL), the ability to acquire new rules, knowledge, and procedures from the input without any parameter update~\cite{brown2020language,dong2024survey}, is what allows a general-purpose model to follow the conventions of a particular user, team, or workplace rather than its pre-trained priors.
This ability matters most for agents that act.
Today, the same frontier models operate computers as computer-use agents (CUAs) and, increasingly, robots as robot-use agents (RUAs)~\cite{chen2026papav,qin2026agenticrobots}.
In both settings, users teach a task in the most natural way: they show a demonstration or state a rule, and expect the agent to apply it to new data and in new surroundings.

Measuring this ability faithfully is difficult, and existing ICL benchmarks fall short in four ways.
First, most of them present the context as text or static images and evaluate offline, whereas acting agents learn from videos and must carry out what they learned in interactive environments.
Second, their rules and data are often drawn from public sources and may have entered pre-training.
Third, as models grow stronger, many tasks become solvable without reading the context at all~\cite{chen2025truemicl}, so success no longer reflects learning.
Fourth, with a single rule per task, they cannot tell whether an agent follows the given rule or its own prior, nor whether it can transfer the rule beyond the demonstrated situation.

The physical world is even less understood.
Recent RUAs increasingly rely on frontier models to learn from in-context demonstrations, yet this ability has been examined only within individual systems, with conflicting conclusions~\cite{guo2026robodawn,zhang2026astra,liu2026roboicl,cheng2026gptpolicy}.
Moreover, although computer use and robot use share the same perceive--act loop, no benchmark asks whether a rule-learning ability that works on screens carries over to robots when the agent and its interface are held fixed.

We present \ours (\cref{fig:teaser}), a benchmark that evaluates the in-context rule learning of agents from the digital to the physical world.
\ours contains 160 tasks in two domains: computer use on web and OS platforms, and tabletop manipulation, in which a dual-arm robot learns from real-world human demonstrations and acts in physics simulation.
Each task asks the agent to reach a goal under a customized, often counter-intuitive rule that is given only in the context, either as a text description or as a video demonstration.
\ours evaluates in-context learning end to end, from inferring the rule to completing the task under it.
\ours is built on four principles:
\begin{itemize}[nosep,leftmargin=*]
  \item \textbf{Leakage-free.} We build all 30 environments and record all 630 demonstration videos (10+ hours) ourselves, around customized rules rather than public material.
  \item \textbf{Context-dependent.} Without the rule, a task can only be completed by chance, and each task comes in three rule versions that demand different behavior in the same situation.
  \item \textbf{Challenging.} Tasks range from short rule applications to long-horizon workflows and fine manipulation, on par with or harder than current frontier benchmarks.
  \item \textbf{Diagnostic.} We test whether learned rules generalize to new instances and unseen environments, and probe the gap between rule recognition and task completion.
\end{itemize}
All agents interact through \harness, a unified harness that presents web applications, desktops, and robot environments through the same kind of graphical interface.
This removes variations from domain-specific interfaces and enables a consistent evaluation across both worlds.

We evaluate six frontier models in their native agent CLIs.
The strongest, Claude Opus 5.5 and GPT-6 Astra, succeed on 82\% and 76\% of computer-use runs but only 27\% and 26\% of robot-use runs, and the other four models trail far behind in both domains.
Without the rule, success drops to at most 8\%, confirming that the tasks require learning from context.
GPT-6 Astra succeeds under all three rule versions on only 38\% of computer-use tasks, half its rate under a single rule.
When asked about the rule before acting, agents usually answer correctly yet fail to carry it out, especially with a robot.
Frontier agents learn rules comparably well from video and text and transfer them across environments, from real-world demonstrations to simulation and to unseen applications, whereas weaker agents rely on text and lose most of their video-based success on new instances.

In summary, our contributions are:
\begin{itemize}[nosep,leftmargin=*]
  \item \ours, a leakage-free benchmark of 160 tasks with three rule versions each, spanning computer use and robot use, built on 30 self-built environments and 630 self-recorded demonstrations.
  \item \harness, a unified harness that turns every world into a user interface, and a diagnostic protocol that together evaluate the same agents in the digital and the physical world across rule versions, generalization levels, rule modalities, and difficulty.
  \item A systematic evaluation of six frontier agents, revealing a large gap between the digital and the physical world and between understanding a rule and executing it.
\end{itemize}

\section{Related Work}
\label{sec:related}

\paragraph{In-context and context learning.}
ICL, popularized by GPT-3~\cite{brown2020language}, has been studied extensively~\cite{dong2024survey}, from what makes demonstrations effective~\cite{min2022rethinking} to its mechanisms~\cite{garg2022can,olsson2022context}.
Other work asks whether context can override priors, \eg with flipped labels~\cite{wei2023larger}, counterfactual task variants~\cite{wu2024reasoning}, or judgments of when context should govern the answer~\cite{li2026jicl}.
Benchmarks evaluate learning a language from a grammar book~\cite{tanzer2024benchmark} and \emph{context learning} of expert-written knowledge and rules~\cite{dou2026clbench,dou2026clbenchlife,li2026learn2play}.
Multimodal ICL is evaluated with interleaved images~\cite{alayrac2022flamingo,zong2025vlicl,chen2025truemicl} and multimodal context learning of visual rules and procedures~\cite{chen2026mmclbench,wei2026clbenchv}.
Video contexts are studied for procedural QA~\cite{dong2026demoicl}, while LMAct~\cite{ruoss2025lmact} and V-ICAL Bench~\cite{fan2026vical} evaluate in-context imitation of multimodal agents in games.
\ours brings rule-centric, leakage-free ICL evaluation to interactive computer and robot environments, with text and video rules for the same task.

\paragraph{Computer-use agents.}
CUAs are evaluated on web~\cite{deng2023mind2web,zhou2024webarena,koh2024visualwebarena}, desktop~\cite{xie2024osworld,bonatti2024windows}, mobile~\cite{rawles2024androidworld}, and grounding~\cite{li2025screenspotpro} benchmarks.
As frontier models exceed 84\% on OSWorld-Verified~\cite{moonshot2026kimik3,qwen2026qwen38}, attention has shifted to long-horizon workflows such as OSWorld~2.0~\cite{yuan2026osworld2}.
Agents build on proprietary or specialized GUI models~\cite{anthropic2024computeruse,openai2025cua,hong2024cogagent,qin2025uitars,agashe2025agents,wang2025opencua}.
Several works let CUAs learn from demonstrations or tutorial videos at inference time~\cite{song2026watch,liu2025onlinevideos,li2025instruction}, and benchmarks evaluate video-guided GUI automation~\cite{lin2024videogui,jang2025videowebarena,fan2026vgguibench}, few-shot demonstrations~\cite{liu2025learnact}, demonstration-guided office workflows~\cite{ding2026uimate}, and user-specific preferences~\cite{jang2026mypcbench}.
None of them varies the rule of the same task or tests it in unseen environments.

\paragraph{Robot-use agents.}
Robot manipulation is dominated by vision-language-action (VLA) policies~\cite{zitkovich2023rt2,kim2024openvla,black2024pi0,pi2025pi05,team2025geminirobotics}, evaluated on LIBERO~\cite{liu2023libero}, RoboTwin~2.0~\cite{chen2025robotwin2}, SimplerEnv~\cite{li2024simpler}, CALVIN~\cite{mees2022calvin}, ManiSkill3~\cite{tao2024maniskill3}, and RoboDojo~\cite{chen2026robodojo}; scores on saturated benchmarks collapse under perturbations~\cite{zhou2025liberopro}.
LLMs and VLMs were first used as planners or code generators~\cite{ahn2022saycan,liang2023code,huang2023voxposer}; more recently, frontier agents act directly as robot-use agents (RUAs) through harnesses that expose robot control as tools~\cite{chen2026showharness,hu2026via,guo2026robodawn,zhang2026astra,yuan2026democratizing,fu2026capx,li2026embodiedevo}, and RUA benchmarks evaluate them across tasks and embodiments~\cite{yang2026robotworld,diao2026liberoagent,sensenova2026astra}.
ICL for robots has been explored with trained in-context policies, including VIMA's multimodal prompts that specify a task~\cite{jiang2023vima,fu2024icrt,vosylius2024instant}, with LLMs reading demonstrations as tokens~\cite{dipalo2024keypoint,yin2024roboprompt}, and by imitation from human videos~\cite{wang2023mimicplay,jain2024vid2robot}.
For frontier RUAs, in-context demonstrations improve success in RoboDawn~\cite{guo2026robodawn}, RoboICL~\cite{liu2026roboicl}, GPT-Policy~\cite{cheng2026gptpolicy}, and VIA~\cite{hu2026via}, while one-shot demonstrations bring no aggregate gain when GPT-6 Astra acts as a policy directly~\cite{zhang2026astra}.
These are method-level ablations with one or a few models.
The concurrent RoboARCade~\cite{sharma2026roboarcade} adds an ICL arm on five tasks, with the agent's own rollout as context.
Closest to our generalization levels, the Imitator Game~\cite{zhou2026imitatorgame} benchmarks robot imitation of human videos at four levels that widen the gap between the demonstrated and the robot's scene.
It and WatchAct~\cite{li2026watchact} evaluate whether a policy reproduces a demonstrated behavior; \ours instead asks whether a general-purpose agent changes its behavior in the same scene according to the rule it is shown, contrasts three rules for the same task, and tests the same agents in the digital world as well.

\paragraph{Unifying digital and physical agents.}
Generalist models are trained on both GUI and embodied data~\cite{reed2022gato,szot2024gea,yang2025magma}, and VisualAgentBench~\cite{liu2024vab} and Embodied Web Agents~\cite{hong2025embodied} evaluate both domains.
Show-Harness~\cite{chen2026showharness} and VIA~\cite{hu2026via} let computer-use agents operate robots through GUI-like interfaces, and robot harnesses reuse coding-agent CLIs as runtimes~\cite{sharma2026roboarcade,yuan2026democratizing}.
PAPAV~\cite{chen2026papav} compares computer use and robot use under one capability framework, and harness design has been surveyed for general agents~\cite{guo2026harnesssurvey}.
None of these studies ICL, which \ours evaluates for the same agents in both domains under a single harness.
\cref{tab:bench_compare} summarizes the differences.

\begin{table}[t]
  \centering
  \setlength{\tabcolsep}{2.2pt}
  \caption{\textbf{Comparison with representative benchmarks.}
  \emph{Evaluated}: what is evaluated (Policy: trained VLA / world-action policies).
  \emph{Env.}: online interactive environment where the model acts and is checked by execution, or offline evaluation on static inputs.
  \emph{Self-built}: data, rules, or environments created by the authors rather than taken from public sources.
  \emph{Modality}: model input, including context and observations (\mT: text, \mI: image, \mV: video).
  \emph{Rule ver.}: multiple rule versions of the same task (\pmark: altered rules, but one version per task).
  \emph{Cross-env.}: the same rule or skill is tested in environments unseen in its demonstration or training.
  \cmark: yes; \pmark: partially; \xmark: no.}
  \label{tab:bench_compare}
  \adjustbox{max width=\columnwidth}{%
  \begin{tabular}{@{}llccccc@{}}
    \toprule
    Benchmark & Evaluated & Env. & Self-built & Modality & Rule ver. & Cross-env. \\
    \midrule
    \multicolumn{7}{@{}l}{\emph{In-context / context learning}} \\
    CL-bench~\cite{dou2026clbench}          & \tLLM  & Offline & \cmark & \mT           & \xmark & \xmark \\
    CL-bench Life~\cite{dou2026clbenchlife} & \tLLM  & Offline & \cmark & \mT           & \xmark & \xmark \\
    VL-ICL Bench~\cite{zong2025vlicl}       & \tMLLM & Offline & \pmark & \mT\,\mI      & \xmark & \xmark \\
    MMCL-Bench~\cite{chen2026mmclbench}     & \tMLLM & Offline & \pmark & \mT\,\mI\,\mV & \pmark & \xmark \\
    CLBench-V~\cite{wei2026clbenchv}        & \tMLLM & Offline & \pmark & \mT\,\mI      & \xmark & \xmark \\
    Demo-ICL~\cite{dong2026demoicl}         & \tMLLM & Offline & \xmark & \mT\,\mV      & \xmark & \xmark \\
    V-ICAL Bench~\cite{fan2026vical}        & \tGame & Online  & \cmark & \mT\,\mI\,\mV & \xmark & \xmark \\
    \midrule
    \multicolumn{7}{@{}l}{\emph{Computer use}} \\
    OSWorld~\cite{xie2024osworld}           & \tCUA  & Online  & \cmark & \mT\,\mI      & \xmark & \xmark \\
    OSWorld 2.0~\cite{yuan2026osworld2}     & \tCUA  & Online  & \cmark & \mT\,\mI      & \xmark & \xmark \\
    AndroidWorld~\cite{rawles2024androidworld} & \tCUA & Online & \cmark & \mT\,\mI     & \xmark & \xmark \\
    OSWorkerBench~\cite{ding2026uimate}     & \tCUA  & Online  & \cmark & \mT\,\mI      & \xmark & \xmark \\
    \midrule
    \multicolumn{7}{@{}l}{\emph{Robot use}} \\
    LIBERO~\cite{liu2023libero}             & \tVLA  & Online  & \cmark & \mT\,\mI      & \xmark & \xmark \\
    RoboTwin 2.0~\cite{chen2025robotwin2}   & \tVLA  & Online  & \cmark & \mT\,\mI      & \xmark & \cmark \\
    RoboDojo~\cite{chen2026robodojo}        & \tVLA\,\tRUA & Online & \cmark & \mT\,\mI & \xmark & \cmark \\
    RobotWorld~\cite{yang2026robotworld}    & \tRUA  & Online  & \pmark & \mT\,\mI      & \xmark & \pmark \\
    Imitator Game~\cite{zhou2026imitatorgame} & \tVLA\,\tRUA & Online & \cmark & \mT\,\mI\,\mV & \xmark & \cmark \\
    \midrule
    Embodied Web Agents~\cite{hong2025embodied} & \tCUA\,\tRUA & Online & \pmark & \mT\,\mI & \xmark & \xmark \\
    \textbf{\ours (ours)}                   & \tCUA\,\tRUA & Online & \cmark & \mT\,\mI\,\mV & \cmark & \cmark \\
    \bottomrule
  \end{tabular}}
\end{table}

\section{The \oursplain Benchmark}
\label{sec:method}

\begin{figure*}[t]
\centering
\includegraphics[width=\linewidth]{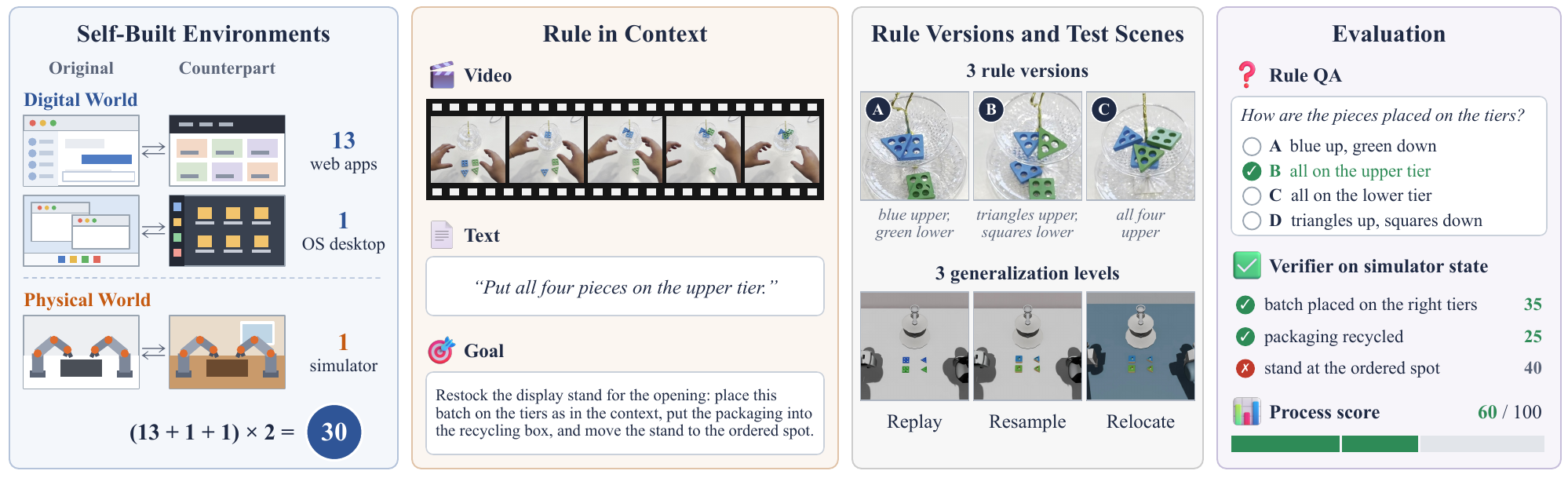}
\caption{\textbf{Building and evaluating \oursplain}, illustrated with a long-horizon robot task: restocking a two-tier display stand with a batch of items, recycling the packaging, and moving the stand to the ordered spot. Every one of the 15 original environments has a counterpart with the same functionality, giving 30 environments (\cref{sec:data}; drawn schematically). The rule is given as a real demonstration video or as text, together with the task goal (\cref{sec:cases}). We show the real final states of the three rule versions and the simulated test scenes at the three generalization levels: \vI replaces the blocks by pastries, and \vE additionally changes the robot arm, the tabletop, and physical parameters. The evaluation panel shows the rule question with its answer key and the task's milestones with their weights; the outcome shown is illustrative (\cref{sec:protocol,sec:metrics}).}
\label{fig:framework}
\end{figure*}

A task in \ours consists of a goal, a rule that changes how the goal must be achieved, an environment in which the agent acts, and a verifier that checks the final state against the rule (\cref{fig:framework}).
We describe how tasks are designed (\cref{sec:cases}), the environments and demonstrations they build on (\cref{sec:data}), the unified harness through which agents act (\cref{sec:harness}), the evaluation protocol (\cref{sec:protocol}), and the metrics (\cref{sec:metrics}).

\subsection{Task Design}
\label{sec:cases}

\paragraph{Rules as the object of learning.}
Every task is built around a rule that changes \emph{how} its goal must be achieved, such as an arbitrary text format, a criterion for selecting items, a filing convention, or the order in which objects are stacked.
We design rules to be customized and often counter-intuitive, so that they cannot be inferred from the goal or from common sense: without the rule, a model can complete the task only by guessing it.
For each task, we design three versions (A/B/C) that share the goal and the initial situation but differ in the rule, \eg rewriting the same notification draft in a different letter case, or boxing a different subset of the same flipped cards (\cref{fig:teaser}); \app{A} shows further examples.
The 160 tasks thus yield 480 task--rule pairs.

\paragraph{Two domains, one taxonomy.}
\ours covers two domains (\cref{fig:teaser}, right).
\emph{Computer use} (110 tasks) takes place in web applications (75 tasks) and an OS desktop (35 tasks), with scenarios ranging from messaging, publishing, finance, coding, and puzzle games to file, document, and system management; \emph{robot use} (50 tasks) consists of tabletop manipulation.
We categorize the tasks of both domains by what their rule specifies:
\emph{selection} (which items to act on), \emph{formatting} (how an output is written, named, or filled in), \emph{labeling and matching} (which label or target each item receives), \emph{ordering} (the order of items or actions), \emph{arrangement} (where objects end up), and \emph{procedure} (the steps or tools used to reach the goal).
Selection and formatting rules dominate computer use, whereas arrangement and ordering rules dominate robot use.

\paragraph{Difficulty.}
We grade each task by its execution demands, independently of its rule:
\begin{itemize}[nosep,leftmargin=*]
  \item \textbf{Computer use} (36/37/37 easy/medium/hard).
  \emph{Easy}: apply the rule to several objects within one application.
  \emph{Medium}: complete a dependent workflow in which the rule is one crucial module.
  \emph{Hard}: complete an end-to-end workflow across applications, in which the rule is again one module.
  \item \textbf{Robot use} (7/19/24).
  \emph{Easy}: a simple manipulation.
  \emph{Medium}: multiple operations or bimanual coordination.
  \emph{Hard}: fine manipulation, or a long-horizon workflow in which the rule is one module.
\end{itemize}
Medium and hard tasks are on par with or harder than the hardest tasks of current computer-use and robot benchmarks~\cite{yuan2026osworld2,chen2026robodojo}.

\subsection{Environments and Demonstrations}
\label{sec:data}

\paragraph{Self-built environments.}
To rule out leakage, we build every environment ourselves: 15 original environments, each paired with a counterpart, 30 in total.
The digital world comprises 13 web applications, from messaging, music, news, and video to travel, shopping, banking, an online judge, and games, and one OS desktop with its own file system and system tools.
Each application maintains realistic state and logic, so a rule must be applied to actual records rather than to a static page.
The physical world is a tabletop simulator built on Isaac Lab, in which a dual-arm robot manipulates objects.
Each counterpart offers the same functionality as its original but changes the environment: a web application or the desktop receives its own layout, navigation, and visual style, and the simulator receives a different robot arm, tabletop, and physical parameters such as friction.
The counterparts allow us to test whether a learned rule transfers to an unseen environment.

\paragraph{Demonstrations.}
For every task and rule version, a human demonstrates the rule, and we write a matching text description.
Computer-use rules are demonstrated as screen recordings in our applications; robot rules are demonstrated by hand in the real world, and each real-world video is paired with a simulated video of the same rule.
A demonstration shows the rule being applied without stating it, so the agent must infer the rule from what it sees.
Every task version is designed by hand and completed by a human through the same \harness interface before release, so it is solvable through that interface and its verifier accepts a correct solution.

\subsection{Unified Harness}
\label{sec:harness}

\begin{figure}[t]
\centering
\includegraphics[width=\linewidth]{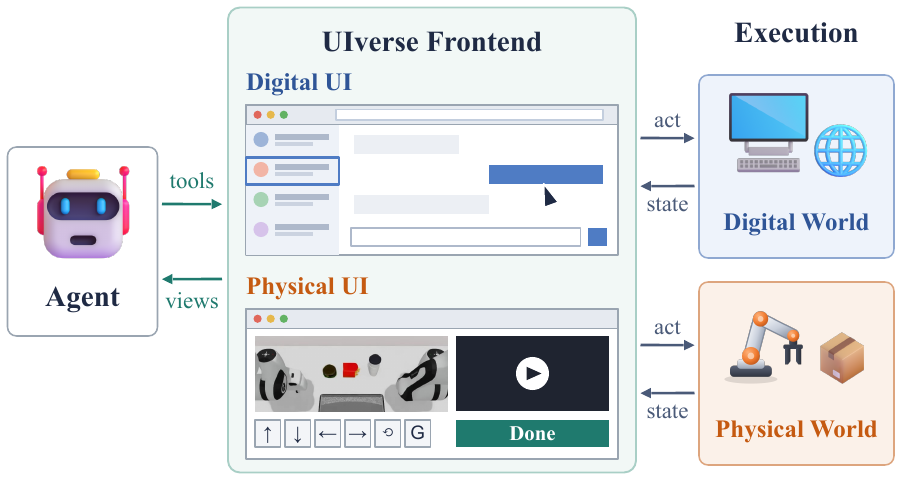}
\caption{\textbf{\harnessplain.} The agent acts only through a frontend that renders every world as a graphical interface: the application's own screen in the digital world, and a control panel with camera views, the demonstration, and arm controls in the physical world. Actions are executed in the corresponding world, which returns its new state.}
\label{fig:harness}
\end{figure}

Computer use and robot use share the same perceive--act loop.
Our harness, \harness, exploits this commonality by turning every world into a user interface (\cref{fig:harness}): a single frontend renders every environment, whether a web application, the OS desktop, or the robot simulator, as a graphical interface in a browser, and offers the agent the same small set of tools to observe the screen, act on it, view the demonstration, and declare completion.
\harness thus unifies the interface between agent and world, \ie observations and actions, while each model keeps its own agent framework.
On the robot interface, the agent sees the head and wrist camera views and operates the motion and gripper controls of each arm; on web and OS interfaces, it clicks, types, and scrolls.
Each agent runs in its own command-line agent with the built-in tools disabled, so it interacts with the world only through the frontend, while verifiers and answer keys remain hidden; \app{B} lists the tools and interfaces.
With the same model and the same interface in both domains, differences between the domains reflect the tasks rather than the interface.
Acting through such an interface is itself part of what we evaluate: just as GUI agents remain necessary although many services offer APIs, an agent serving users across worlds must act through the interfaces these worlds present.

\subsection{Evaluation Protocol}
\label{sec:protocol}

\paragraph{Rule versions.}
Since the three versions of a task demand different behavior in the same situation, an agent that acts on a fixed habit or prior can satisfy at most one of them; consistent success across versions requires following the given rule.

\paragraph{Generalization levels.}
We distinguish three levels.
\vR restores the demonstrated situation.
\vI samples a new instance of the task independently of the demonstration, from the same distribution and in the same environment (IID).
\vE keeps the goal and the rule but relocates the task to the counterpart environment.
Every task supports \vE (\cref{tab:protocol}).
\vR is defined only where the demonstration shows the whole task: easy computer-use tasks and most robot tasks, whereas medium and hard computer-use demonstrations teach only the rule-bearing module of a longer workflow.
Robot tasks that hinge on unique objects, such as a xylophone or screws, admit no independent instance and thus no \vI.

\begin{table}[H]
  \centering
  \footnotesize
  \setlength{\tabcolsep}{5pt}
  \caption{\textbf{Tasks supporting each generalization level} among all 160 tasks, listed as \vR\,/\,\vI\,/\,\vE.}
  \label{tab:protocol}
  \begin{tabular}{@{}lll@{}}
    \toprule
    Difficulty & Computer use & Robot use \\
    \midrule
    Easy   & 36 / 36 / 36 & 7 / 7 / 7 \\
    Medium & 0 / 37 / 37  & 19 / 19 / 19 \\
    Hard   & 0 / 37 / 37  & 19 / 16 / 24 \\
    \bottomrule
  \end{tabular}
\end{table}

\begin{table*}[t]
  \centering
  \footnotesize
  \setlength{\tabcolsep}{4.5pt}
  \caption{\textbf{Main results} (success rate, \%) on all 160 tasks with rule version C. Each cell reports \tv{text}{video}: the rule is given as a text description (shaded) or as a video demonstration. Overall: mean of the two domain averages; Avg.: all tasks of the domain. Computer-use tasks are evaluated at \vI; robot tasks at \vI, or at \vR for the 8 tasks that admit no independent instance, on which no model succeeds. Models are sorted by overall score with text rules. Best text and video results per column in bold.}
  \label{tab:main}
  \begin{tabular}{@{}l c cccc cccc@{}}
    \toprule
    & & \multicolumn{4}{c}{Computer use (110)} & \multicolumn{4}{c}{Robot use (50)} \\
    \cmidrule(lr){3-6}\cmidrule(lr){7-10}
    Model & Overall & Avg. & Easy & Medium & Hard & Avg. & Easy & Medium & Hard \\
    \midrule
    Claude Opus 5.5 & \tv{\textbf{53.9}}{\textbf{54.9}} & \tv{\textbf{81.8}}{\textbf{81.8}} & \tv{\textbf{97.2}}{\textbf{94.4}} & \tv{\textbf{89.2}}{\textbf{86.5}} & \tv{\textbf{59.5}}{\textbf{64.9}} & \tv{\textbf{26.0}}{\textbf{28.0}} & \tv{28.6}{\textbf{57.1}} & \tv{\textbf{57.9}}{42.1} & \tv{0.0}{\textbf{8.3}} \\
    GPT-6 Astra & \tv{49.7}{52.6} & \tv{75.5}{77.3} & \tv{88.9}{88.9} & \tv{86.5}{81.1} & \tv{51.4}{62.2} & \tv{24.0}{\textbf{28.0}} & \tv{\textbf{57.1}}{\textbf{57.1}} & \tv{31.6}{\textbf{47.4}} & \tv{\textbf{8.3}}{4.2} \\
    Qwen3.8-Max & \tv{19.8}{17.5} & \tv{33.6}{29.1} & \tv{63.9}{63.9} & \tv{29.7}{18.9} & \tv{8.1}{5.4} & \tv{6.0}{6.0} & \tv{0.0}{0.0} & \tv{15.8}{15.8} & \tv{0.0}{0.0} \\
    DeepSeek-V4.1-Flash & \tv{17.8}{9.3} & \tv{23.6}{14.5} & \tv{55.6}{38.9} & \tv{13.5}{5.4} & \tv{2.7}{0.0} & \tv{12.0}{4.0} & \tv{14.3}{0.0} & \tv{26.3}{10.5} & \tv{0.0}{0.0} \\
    GLM-5.3-Flash & \tv{17.7}{15.1} & \tv{35.5}{28.2} & \tv{80.6}{63.9} & \tv{18.9}{16.2} & \tv{8.1}{5.4} & \tv{0.0}{2.0} & \tv{0.0}{0.0} & \tv{0.0}{5.3} & \tv{0.0}{0.0} \\
    Kimi K3 & \tv{14.1}{12.3} & \tv{28.2}{24.5} & \tv{63.9}{55.6} & \tv{16.2}{10.8} & \tv{5.4}{8.1} & \tv{0.0}{0.0} & \tv{0.0}{0.0} & \tv{0.0}{0.0} & \tv{0.0}{0.0} \\
    \bottomrule
  \end{tabular}
\end{table*}

\paragraph{Rule modality.}
The rule is given either as a text description or as a video demonstration, which lets us compare learning from language and from observation.
The agent inspects a video through the interface, viewing any frame or several chosen frames at once as often as it needs, so every model receives the video in the same way.

\paragraph{Diagnoses.}
To separate understanding a rule from executing it, every task comes with a multiple-choice question about its rule.
It is posed in a separate branch of the agent's context, created after the agent has inspected the demonstration and before its first action, so the question and the run share the same context, while the answer never reaches the execution run.
To confirm that a task requires ICL, we also run it with the task goal alone and no rule.

\subsection{Metrics}
\label{sec:metrics}

\paragraph{Success and process score.}
When the agent declares completion or exhausts its budget, a verifier checks the final state of the environment against the rule version given in the context.
A run $j$ succeeds, $s_j=1$, only if the goal is reached under that rule.
Since binary success hides partial progress, the verifier also assigns a process score $p_j\in[0,1]$, with $p_j=1$ if $s_j=1$, which gives partial credit for the rule-bearing steps that a failed run completes correctly.
The verifiers and the scoring protocol of each domain are detailed in \app{D}.

\paragraph{Aggregation.}
For a domain $d\in\{\text{CUA},\text{RUA}\}$ with task set $\mathcal{T}_d$, the domain score is the mean success
\begin{equation}
  S_d=\frac{1}{|\mathcal{T}_d|}\sum_{t\in\mathcal{T}_d} s_t ,
\end{equation}
and the overall score weights the two domains equally,
\begin{equation}
  S=\tfrac{1}{2}\left(S_{\text{CUA}}+S_{\text{RUA}}\right),
\end{equation}
so that the larger computer-use domain does not dominate.
Scores at a given difficulty or generalization level average over the corresponding tasks, and the same aggregation applies to process scores.

\paragraph{Text--video gap.}
To compare learning from language and from observation, we evaluate every task with both rule modalities and define the text--video gap of a model on a task set $\mathcal{T}$ as
\begin{equation}
  \Delta_{\mathrm{tv}}=\frac{1}{|\mathcal{T}|}\sum_{t\in\mathcal{T}}\bigl(s_t^{\text{text}}-s_t^{\text{video}}\bigr),
\end{equation}
where $s_t^{\text{text}}$ and $s_t^{\text{video}}$ are the outcomes of the same task under the two modalities.
A positive gap means that the model learns the rule better from its description than from its demonstration.
Because the two runs share the task, we test whether the gap differs from zero with a two-sided sign test over the tasks on which exactly one modality succeeds, report 95\% bootstrap confidence intervals over tasks, and correct $p$-values with the Holm method within each family of comparisons, where a family contains the comparisons along one factor: models, rule types, difficulty levels, or domains.

\section{Experiments}
\label{sec:experiments}

\subsection{Setup}
\label{sec:setup}

\begin{table*}[t]
  \centering
  \footnotesize
  \setlength{\tabcolsep}{4pt}
  \caption{\textbf{Generalization} (success rate, \%, \tv{text}{video}) with rule version C on the 54-task subset (\app{C}); sizes in parentheses. Computer use is paired: \vR vs.\ \vI on the same 13 tasks, and \vI vs.\ \vE on the same 37 tasks. In robot use, each of the 17 subset tasks is evaluated at every level it supports.}
  \label{tab:generalization}
  \begin{tabular}{@{}l cc cc ccc@{}}
    \toprule
    & \multicolumn{4}{c}{Computer use} & \multicolumn{3}{c}{Robot use} \\
    \cmidrule(lr){2-5}\cmidrule(l){6-8}
    Model & \vR (13) & \vI (13) & \vI (37) & \vE (37) & \vR (14) & \vI (15) & \vE (17) \\
    \midrule
    Claude Opus 5.5 & \tv{84.6}{100.0} & \tv{92.3}{92.3} & \tv{78.4}{78.4} & \tv{70.3}{81.1} & \tv{21.4}{35.7} & \tv{33.3}{26.7} & \tv{29.4}{11.8} \\
    GPT-6 Astra & \tv{76.9}{100.0} & \tv{76.9}{84.6} & \tv{67.6}{75.7} & \tv{70.3}{73.0} & \tv{35.7}{28.6} & \tv{13.3}{20.0} & \tv{5.9}{17.6} \\
    Qwen3.8-Max & \tv{69.2}{100.0} & \tv{69.2}{69.2} & \tv{27.0}{32.4} & \tv{37.8}{37.8} & \tv{7.1}{0.0} & \tv{0.0}{6.7} & \tv{5.9}{0.0} \\
    DeepSeek-V4.1-Flash & \tv{53.8}{69.2} & \tv{53.8}{15.4} & \tv{24.3}{8.1} & \tv{32.4}{18.9} & \tv{7.1}{7.1} & \tv{20.0}{6.7} & \tv{17.6}{0.0} \\
    GLM-5.3-Flash & \tv{53.8}{84.6} & \tv{46.2}{53.8} & \tv{21.6}{21.6} & \tv{32.4}{35.1} & \tv{7.1}{0.0} & \tv{0.0}{6.7} & \tv{5.9}{5.9} \\
    Kimi K3 & \tv{53.8}{84.6} & \tv{46.2}{46.2} & \tv{18.9}{16.2} & \tv{35.1}{35.1} & \tv{0.0}{0.0} & \tv{0.0}{0.0} & \tv{0.0}{0.0} \\
    \bottomrule
  \end{tabular}
\end{table*}

\begin{figure*}[t]
\centering
\includegraphics[width=\linewidth]{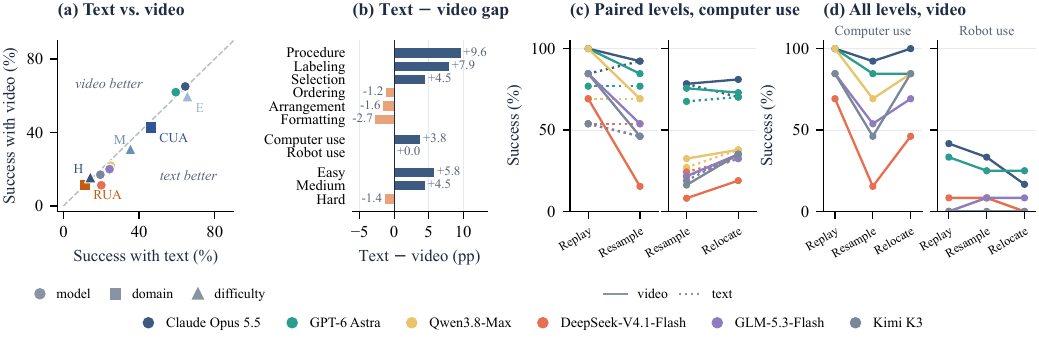}
\caption{\textbf{Analysis} (rule version C). (a) Text vs.\ video success. (b) Text $-$ video gap by rule type, domain, and difficulty. (c) Computer-use success at paired levels. (d) Video success at all three levels, on the tasks supporting all of them.}

\label{fig:analysis}
\end{figure*}

\paragraph{Models.}
We evaluate six frontier models, each in its native agent CLI as it is deployed in practice, except Qwen3.8-Max, which runs in Claude Code:
Claude Opus 5.5~\cite{anthropic2026opus55} in Claude Code,
GPT-6 Astra~\cite{openai2026astra} in Codex,
Qwen3.8-Max~\cite{qwen2026qwen38} in Claude Code,
GLM-5.3-Flash~\cite{zhipu2026glm53flash} in ZCode,
Kimi K3~\cite{moonshot2026kimik3} in Kimi Code,
and DeepSeek-V4.1-Flash~\cite{deepseek2026v41flash} in DeepSeek Harness (DSH).
All use high reasoning effort.
Since all our findings compare conditions within a model, each comparison keeps the model's agent fixed.

\paragraph{Protocol.}
The main experiment evaluates all six models on all 160 tasks with rule version C, with both the text rule and the video demonstration.
The ablations evaluate three models that span the range of the main results; the tasks of each experiment are given in the captions and \app{C}, and process scores and per-task results in \app{E}.

\subsection{Main Results}
\label{sec:main_results}

\paragraph{The physical world is much harder.}
\textbf{Agents that master in-context rules on screens largely fail to apply them with a robot.}
In \cref{tab:main}, Claude Opus 5.5 and GPT-6 Astra reach overall scores of 50--55\%, and the other four models 9--20\%.
The strongest agents solve 76--82\% of computer-use tasks but only 24--28\% of robot-use tasks, and no other model exceeds 12\% in robot use.
With the model, interface, and rule design shared, the gap lies in applying a learned rule in the physical world, where it must be grounded in camera views and carried out through long sequences of precise arm motions; \cref{sec:ablation} separates these parts, and process scores show the same gap (\app{E}).

\paragraph{Difficulty.}
\textbf{Success falls with difficulty in computer use, and hard tasks are the hardest in both domains.}
In computer use, success decreases from easy to hard for every model and modality: every model solves about 39\% or more of easy tasks, while on medium and hard ones only Claude Opus 5.5 and GPT-6 Astra exceed 30\%.
In robot use, hard tasks, which require fine manipulation or long workflows, are the hardest for every model, with no model above 8.3\%, whereas easy and medium robot tasks are solved at similar rates.

\subsection{Analysis}
\label{sec:analysis}

\subsubsection{Text vs.\ Video}

\textbf{Frontier agents learn rules comparably well from video and from text, whereas weaker agents rely on text.}
In computer use, the text--video gap $\Delta_{\mathrm{tv}}$ of Claude Opus 5.5 and GPT-6 Astra lies within two points of zero (\cref{fig:analysis}a).
The other four models all do better with text, by 6.1 points on average ($p=0.02$ after correction) and by up to 9.1 points for DeepSeek-V4.1-Flash.
The same holds in robot use, where their gaps are $-2.0$ and $-4.0$ points.

\textbf{Video is harder to decode when the rule is a hidden criterion.}
Rules that are visible in the final state, such as a format, an arrangement, or an order, are learned equally well from both modalities (\cref{fig:analysis}b).
Rules that act as a hidden criterion, \ie which procedure to follow, which label each item receives, or which items to select, show the largest text advantage, from 4.5 to 9.6 points, most clearly for procedure rules.
On hard computer-use tasks, the frontier agents even do better with video on formatting rules, where a demonstration shows the exact target of a long workflow.

\subsubsection{Generalization}

\textbf{Evaluating only in the demonstrated situation overestimates learning from video.}
In the demonstrated situation, an agent can succeed by repeating what it saw without learning the rule.
On the 13 computer-use tasks with \vR (\cref{tab:generalization}, \cref{fig:analysis}c), text-based success is the same at \vR and \vI, but video-based success falls by 29.5 points once the instance is resampled ($p<0.001$).
The fall comes almost entirely from the weaker models (84.6\% to 46.2\%), whose success at \vR thus largely reflects repeating the demonstrated actions, whereas the frontier agents stay above 85\% at both levels.
Testing only in the demonstrated situation would thus credit imitation as rule learning.

\textbf{Frontier agents transfer learned rules across environments.}
In computer use, they keep their success when a task moves to a counterpart application with a different layout, navigation, and visual style, staying within three points of \vI on the 37 tasks with \vE; the weaker models even gain, suggesting that some counterparts are easier to navigate.
In robot use, every run already transfers a rule from a real-world demonstration to a simulated scene, and this domain gap costs little: a simulated demonstration that matches the execution domain adds at most one task (\cref{tab:abl_norule}, \emph{Sim}).
\vE further changes the robot arm, the tabletop, and physical parameters, which makes execution itself harder and plausibly explains the lower robot success at \vE (\cref{fig:analysis}d).

\subsection{Ablation Studies}
\label{sec:ablation}

\begin{table}[t]
  \centering
  \scriptsize
  \setlength{\tabcolsep}{3pt}
  \caption{\textbf{Ablations on the 54-task subset} (success rate, \%; 37 computer-use and 17 robot tasks). Top: rule input with version C; \emph{None}: goal only; robot \emph{Video}/\emph{Sim}: real-world/simulated demonstration. Bottom: rule versions with video, each judged by its own rule; \emph{All}: solved under all three.}
  \label{tab:abl_norule}
  \begin{tabular}{@{}l cccc cccc@{}}
    \toprule
    & \multicolumn{4}{c}{Computer use} & \multicolumn{4}{c}{Robot use} \\
    \cmidrule(lr){2-5}\cmidrule(lr){6-9}
    Model & None & Text & Video & & None & Text & Video & Sim \\
    \midrule
    GPT-6 Astra & 5.4 & 67.6 & 75.7 & & 0.0 & 11.8 & 17.6 & 23.5 \\
    DeepSeek-V4.1-Flash & 2.7 & 24.3 & 8.1 & & 0.0 & 17.6 & 5.9 & 5.9 \\
    GLM-5.3-Flash & 8.1 & 21.6 & 21.6 & & 0.0 & 0.0 & 5.9 & 5.9 \\
    \midrule
    & A & B & C & All & A & B & C & All \\
    \midrule
    GPT-6 Astra & 48.6 & 48.6 & 75.7 & 37.8 & 29.4 & 11.8 & 17.6 & 11.8 \\
    DeepSeek-V4.1-Flash & 29.7 & 24.3 & 8.1 & 8.1 & 11.8 & 0.0 & 5.9 & 0.0 \\
    GLM-5.3-Flash & 29.7 & 27.0 & 21.6 & 16.2 & 0.0 & 11.8 & 5.9 & 0.0 \\
    \bottomrule
  \end{tabular}
\end{table}

\begin{figure}[t]
\centering
\includegraphics[width=\linewidth]{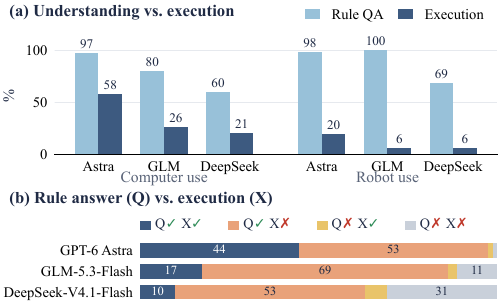}
\caption{\textbf{Understanding vs.\ execution} on the 54-task subset with video rules, pooled over versions A/B/C. (a) Accuracy of the rule question answered before acting and success of the matching run. (b) Share of (task, version) pairs by answer correctness (Q) and execution success (X).}
\label{fig:qa}
\end{figure}

\paragraph{No rule.}
\textbf{Without the rule, the tasks are practically unsolvable.}
Success falls to 2.7--8.1\% on computer use and to 0\% on robot use (\cref{tab:abl_norule}), and the few successes are lucky guesses in which the rule coincides with a cue in the environment (\app{F}).

\paragraph{Rule versions.}
\textbf{Few tasks are solved under all three rules.}
GPT-6 Astra solves 48.6\%, 48.6\%, and 75.7\% of computer-use tasks under versions A, B, and C, but only 37.8\% under all three (\cref{tab:abl_norule}, bottom); in robot use, no model exceeds 11.8\%.
Part of this drop reflects the stricter criterion, but the differences between versions also point to properties of the rules themselves, such as their agreement with the agent's priors, their counterfactual strength, \ie how far they depart from the behavior the goal alone suggests, and the cues the environment offers for them (\app{F}).
Either way, a single rule per task overestimates how reliably an agent follows a user's rule.

\paragraph{Understanding vs.\ execution.}
\textbf{Agents usually identify the rule correctly but fail to carry it out.}
GPT-6 Astra answers 97.3\% of computer-use and 98.0\% of robot-use multiple-choice rule questions correctly before acting (\cref{fig:qa}), yet succeeds on only 57.7\% and 19.6\% of the corresponding runs; GLM-5.3-Flash answers every robot question correctly but succeeds on 5.9\% of the runs.
Over all (task, version) pairs, the rule is identified but not executed in 53--69\% of the cases, while success after a wrong answer is rare ($\le$6\%).
Identifying the rule is thus far from sufficient for completing the task, especially in the physical world (Appendices E, G, and H).

\section{Conclusion}
\label{sec:conclusion}

We presented \ours, a benchmark that evaluates the in-context learning of agents from the digital to the physical world, with three rule versions per task, text and video rules, and three generalization levels, all under the unified harness \harness.
Across six frontier models, the strongest agents solve most computer-use tasks but fail on about three quarters of robot-use tasks, and they usually identify the rule correctly yet fail to carry it out, which places the main obstacle in execution, above all in the physical world.
Frontier agents learn rules comparably well from video and text and transfer them across environments, whereas weaker agents rely on text and, in the demonstrated situation, succeed by repeating the shown actions.
Few tasks are solved under all three rules, so evaluating a single rule per task overestimates how reliably agents follow what users ask.
We hope that \ours helps measure and close these gaps on the way to agents that truly learn from what users show them.

\section{Limitations}
\label{sec:limitations}

Hand-built environments limit \ours to 160 tasks, ablations use three models, robot tasks run in simulation, and no task yet combines computer and robot use.

\section*{LLM Usage Statement}
We used an LLM-based assistant (Claude, Anthropic) to polish the wording of the manuscript and to help debug code, and we used LLMs to obtain feedback on drafts of the paper.
LLMs also assisted in annotating the dataset, and all annotations were manually verified by the authors.
All research ideas, the benchmark design, the environments, tasks, rules, and demonstrations, the experiments, and the analysis were conceived and carried out by the authors, who also wrote and verified all content of the paper.
The frontier models evaluated in this paper are the subject of the study and were not used as tools for producing it.

\FloatBarrier

{
    \small
    \bibliographystyle{ieeenat_fullname}
    \bibliography{main}
}

\clearpage
\appendix
\renewcommand{\thesection}{\Alph{section}}
\section*{Appendix}
\section{Rule Examples}
\label{app:examples}

\begin{figure*}[t]
  \centering
  \setlength{\tabcolsep}{1.5pt}
  \renewcommand{\arraystretch}{0.6}
  
  \begin{tabular}{@{}ccc@{}}
    \footnotesize Rule A & \footnotesize Rule B & \footnotesize Rule C \\
    \includegraphics[width=0.325\linewidth]{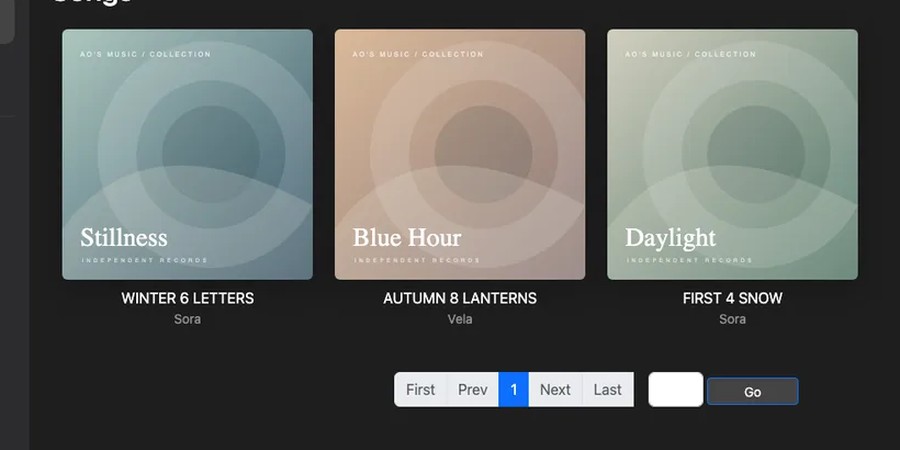} & \includegraphics[width=0.325\linewidth]{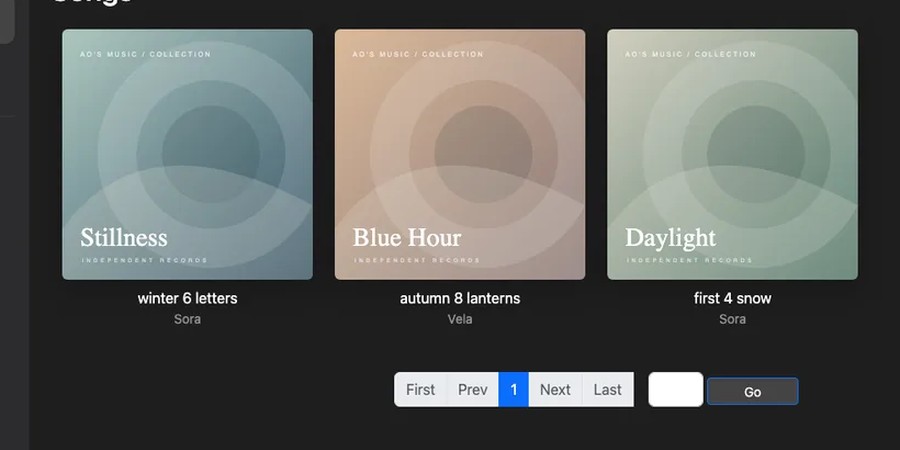} & \includegraphics[width=0.325\linewidth]{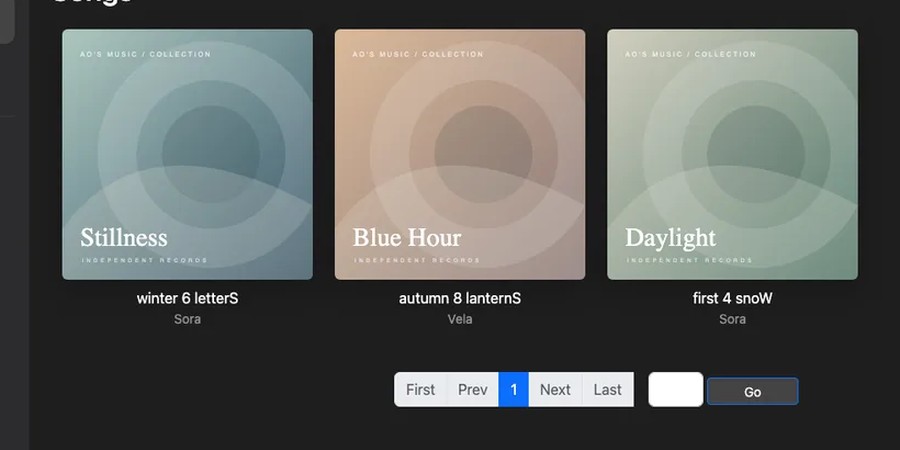} \\
    \footnotesize \textsc{Replay} & \footnotesize \textsc{Resample} & \footnotesize \textsc{Relocate} \\
    \includegraphics[width=0.325\linewidth]{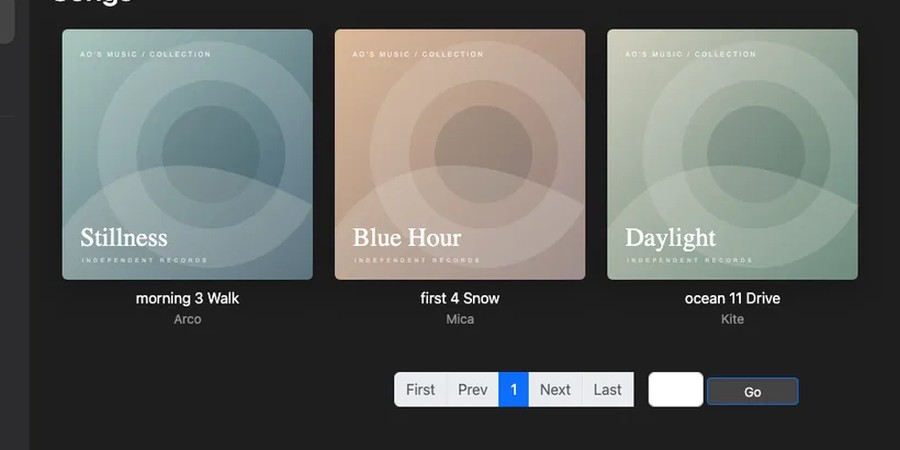} & \includegraphics[width=0.325\linewidth]{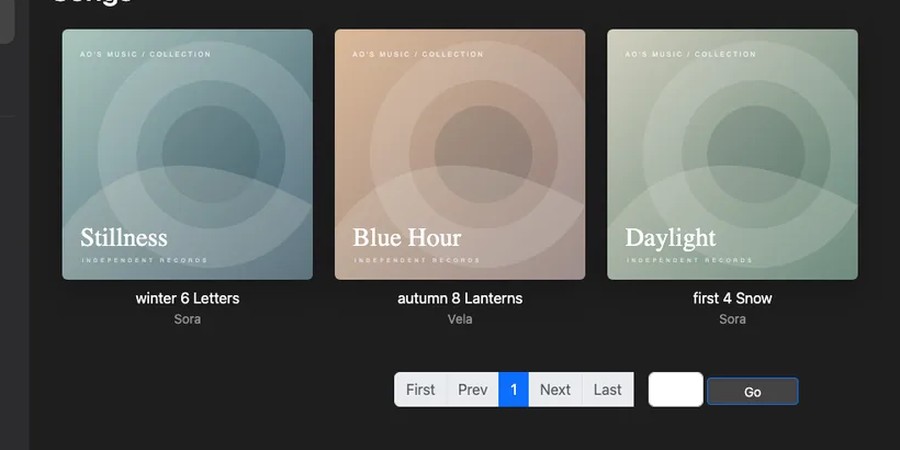} & \includegraphics[width=0.325\linewidth]{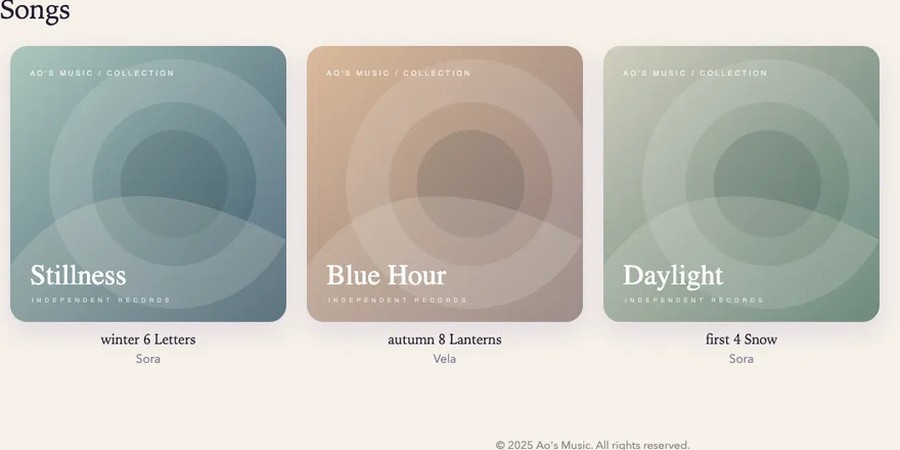} \\[4pt]
    \footnotesize Rule A & \footnotesize Rule B & \footnotesize Rule C \\
    \includegraphics[width=0.325\linewidth]{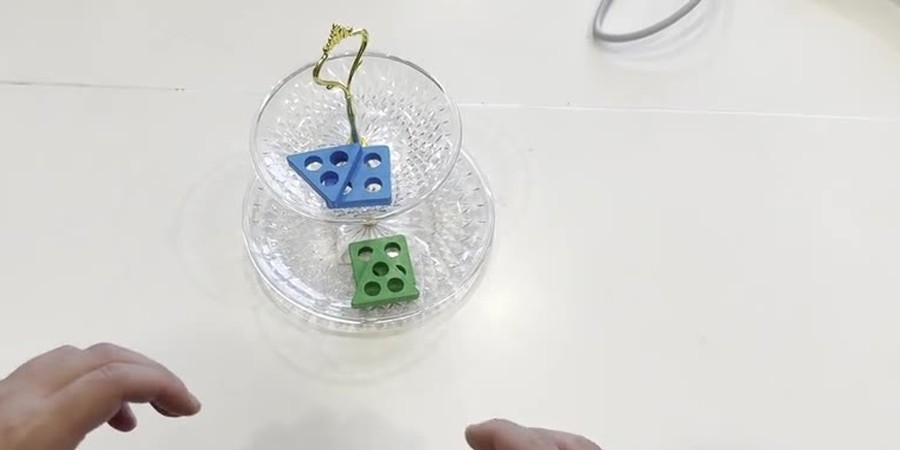} & \includegraphics[width=0.325\linewidth]{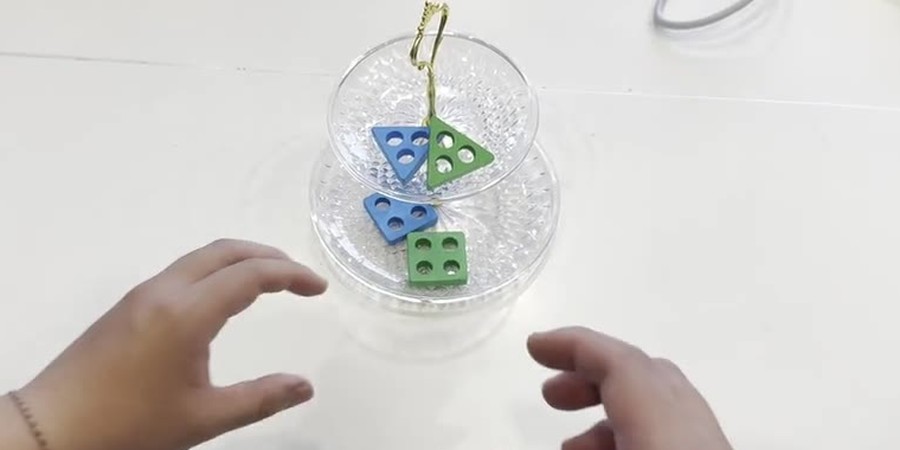} & \includegraphics[width=0.325\linewidth]{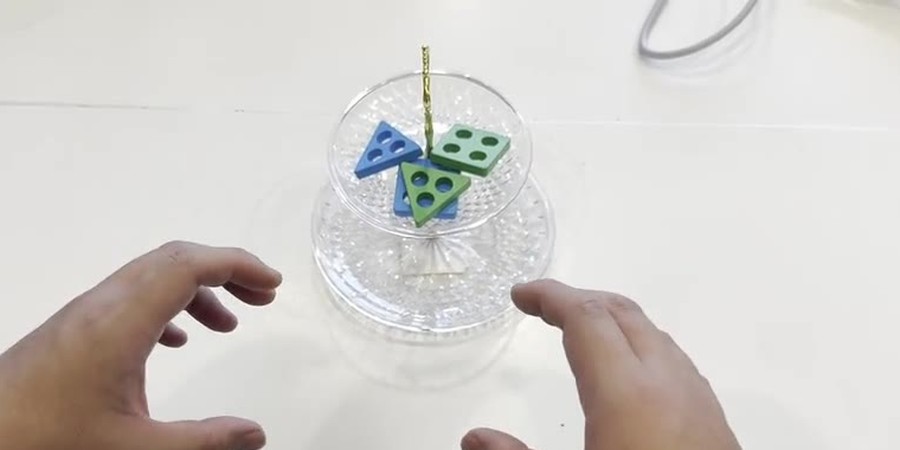} \\
    \footnotesize \textsc{Replay} & \footnotesize \textsc{Resample} & \footnotesize \textsc{Relocate} \\
    \includegraphics[width=0.325\linewidth]{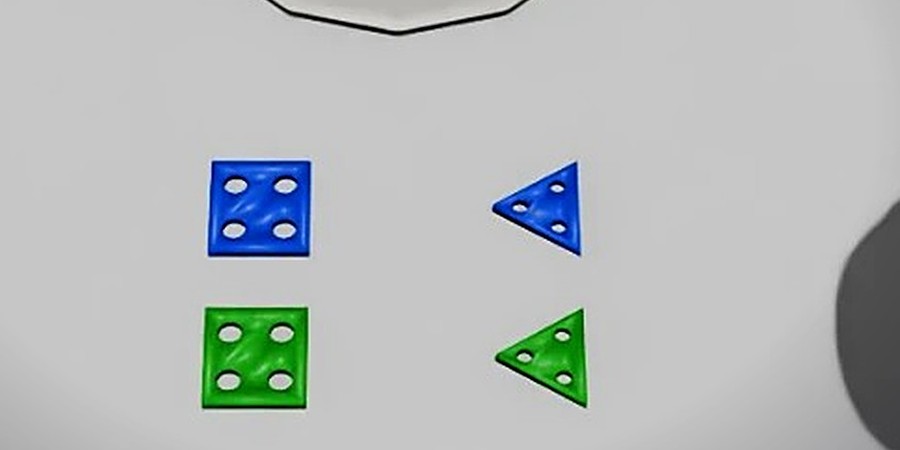} & \includegraphics[width=0.325\linewidth]{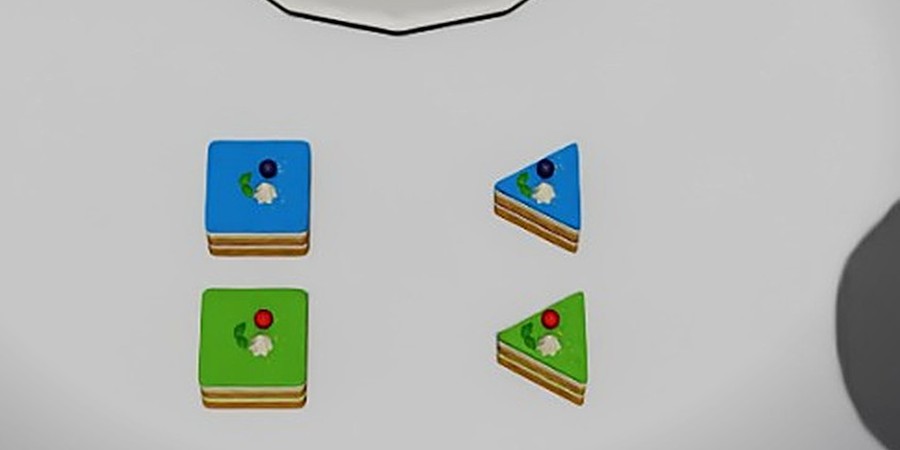} & \includegraphics[width=0.325\linewidth]{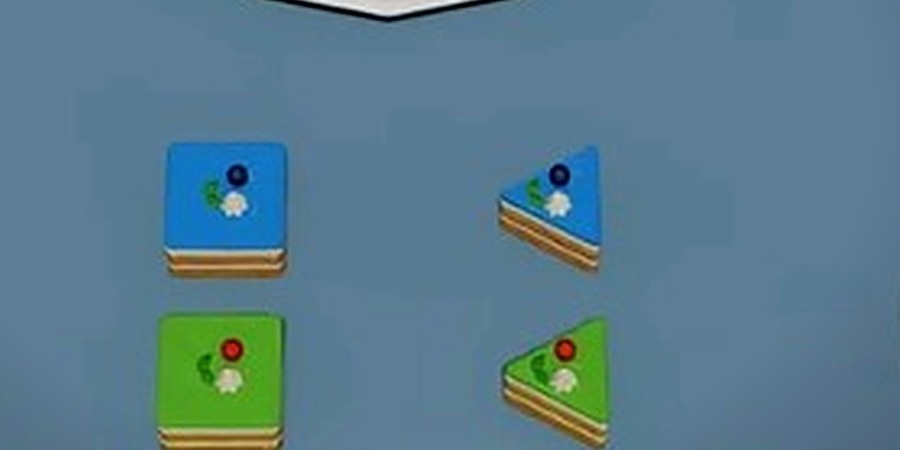} \\
  \end{tabular}
  \caption{\textbf{One goal, three rules, three levels.} Top two rows, computer use: normalize three song titles in a music library. The library after applying rule A (uppercase the title), B (lowercase the title), and C (uppercase only its final letter), and the initial library at the three levels; \textsc{Resample} draws new songs, and \textsc{Relocate} moves the task to a counterpart application with a different layout and style. Bottom two rows, robot use: restock a two-tier display stand (Fig.~\ref{fig:framework}). The end of the real-world demonstration of rule A (blue upper, green lower), B (triangles upper, squares lower), and C (all four upper), and the initial simulated scene at the three levels; \textsc{Resample} replaces the blocks by pastries, and \textsc{Relocate} additionally changes the robot arm, the tabletop, and physical parameters.}
  \label{fig:examples}
\end{figure*}

\cref{fig:examples} shows one computer-use and one robot-use task under its three rule versions and at its three generalization levels, and \cref{tab:examples} lists the rule versions of further tasks.
In every task, the versions share the goal and the initial situation but require different behavior, and the levels change the situation while keeping the goal and the rule.
All 160 tasks with their rule versions, the environments, demonstrations, verifiers, and the \harnessplain harness are publicly available at the links on the first page.

\begin{table}[!htb]
  \centering
  \scriptsize
  \setlength{\tabcolsep}{3pt}
  \caption{\textbf{Rule versions of further tasks.}}
  \label{tab:examples}
  \begin{tabular}{@{}>{\raggedright\arraybackslash}p{0.24\linewidth}>{\raggedright\arraybackslash}p{0.22\linewidth}>{\raggedright\arraybackslash}p{0.22\linewidth}>{\raggedright\arraybackslash}p{0.22\linewidth}@{}}
    \toprule
    Goal & Rule A & Rule B & Rule C \\
    \midrule
    Rewrite three customers' notification drafts and send each back. & Lowercase the first letter, uppercase the rest. & Uppercase the first letter, lowercase the rest. & Title-case every word. \\
    \addlinespace
    Open one safe cell on each Minesweeper board. & Minimize the sum of neighboring clues. & Maximize the unopened neighbors. & Choose the cell closest to the upper-left corner. \\
    \addlinespace
    Submit a revised packet with a marked title. & \texttt{[r8] Title} & \texttt{Title [r8]} & \texttt{r8 - Title} \\

    \midrule
    Set the open/closed state of two drawers. & Open only the upper red drawer. & Open only the lower blue drawer. & Open both drawers. \\
    \addlinespace
    Load four blocks into a 2$\times$2 box in a given cell order. & Red, yellow, blue, green. & Red, green, blue, yellow. & Green, blue, yellow, red. \\
    \addlinespace
    Move three T-blocks into a tray with the demonstrated transformation. & Preserve positions and orientations. & Preserve positions, rotate each block by 90$^\circ$. & Mirror positions, preserve orientations. \\
    \bottomrule
  \end{tabular}
\end{table}

\section{The \harnessplain Harness}
\label{app:harness}

\harnessplain presents every world as a graphical interface and gives the agent the same kinds of tools in every world: observe the interface, act on it, inspect the demonstration, and stop (\cref{tab:tools}).
Each model runs in its own agent CLI with all built-in tools disabled, and the tools of \harnessplain, served over the Model Context Protocol, are its only access to the world; the agent can neither inspect internal state nor read verifiers or answer keys.

\paragraph{Episodes.}
For each episode, the runner binds a task, a rule version and modality, a generalization level, and a seed, and creates an isolated world instance with its own interface and a fresh model session.
The first message gives the goal, the rule as text or as a demonstration video, and a screenshot of the interface.

\paragraph{Digital world.}
The interface is the application's own screen.
The agent acts in screenshot pixel coordinates with mouse and keyboard inputs, as a human user would, and receives the new screenshot.

\paragraph{Physical world.}
The interface is a control panel (Fig.~\ref{fig:harness}) with the head camera and both wrist cameras, buttons that move each arm along its axes, rotate it, and open or close its gripper, fields for the step size and the rotation angle, and \emph{Commit}, \emph{Done}, and \emph{Give up} buttons.
A status area shows the robot's public state, such as the end-effector poses, the gripper openings, and the outcome of the previous actions; object poses are never given, so the agent must locate objects in the camera views.
In each turn, the agent selects a short sequence of panel actions, each an arm button with its step size or angle, and commits them at once.
The simulator executes the actions in order: an inverse-kinematics controller turns each end-effector motion into joint commands, and the physics engine resolves contacts, friction, and grasping.
The agent thus decides where to approach, how far to move, how to orient the gripper, and when to grasp, as a teleoperator would.
After the actions, the simulation pauses, and the agent receives the new panel view with feedback on whether each motion was completed or blocked.

\begin{table}[!htb]
  \centering
  \scriptsize
  \setlength{\tabcolsep}{3pt}
  \caption{\textbf{Tools of \harnessplain} by purpose.}
  \label{tab:tools}
  \begin{tabular}{@{}>{\raggedright\arraybackslash}p{0.17\linewidth}>{\raggedright\arraybackslash}p{0.38\linewidth}>{\raggedright\arraybackslash}p{0.38\linewidth}@{}}
    \toprule
    Purpose & Digital world & Physical world \\
    \midrule
    Observe & Screenshot of the application & Screenshot of the control panel \\
    Act & Click, drag, scroll, type, or press keys at pixel coordinates & Press arm buttons with step sizes and angles, then commit \\
    Demonstration & Frames of the demonstration at chosen times & Frames of the demonstration at chosen times \\
    Stop & Declare completion & Press \emph{Done} or \emph{Give up} \\
    \bottomrule
  \end{tabular}
\end{table}

\section{Task Subset}
\label{app:subset}

The main experiment evaluates all 160 tasks.
The generalization experiment and the ablations use a 54-task subset drawn from every combination of domain and difficulty, about one third of each (\cref{tab:subset}), so that the subset keeps the difficulty mix of each domain; it also keeps the mix of rule types, with about one third of the tasks of every type.
In computer use, \vR is defined for easy tasks, so 13 subset tasks are evaluated at \vR and all 37 at \vI and \vE.
In robot use, each of the 17 tasks is evaluated at every level it supports: 14 at \vR, 15 at \vI, and 17 at \vE.

\begin{table}[!htb]
  \centering
  \scriptsize
  \setlength{\tabcolsep}{3.2pt}
  \caption{\textbf{Tasks in the subset and in the full benchmark} by domain and difficulty (top) and by rule type (bottom; selection, formatting, labeling and matching, ordering, arrangement, procedure).}
  \label{tab:subset}
  \begin{tabular}{@{}l ccc ccc ccc@{}}
    \toprule
    & \multicolumn{3}{c}{Web} & \multicolumn{3}{c}{OS} & \multicolumn{3}{c}{Robot} \\
    \cmidrule(lr){2-4}\cmidrule(lr){5-7}\cmidrule(l){8-10}
    & E & M & H & E & M & H & E & M & H \\
    \midrule
    Subset & 9 & 8 & 8 & 4 & 4 & 4 & 2 & 7 & 8 \\
    All tasks & 25 & 25 & 25 & 11 & 12 & 12 & 7 & 19 & 24 \\
    \midrule
    & \multicolumn{2}{c}{Sel.} & \multicolumn{2}{c}{Fmt.} & \multicolumn{2}{c}{Lab.} & Ord. & Arr. & Proc. \\
    \midrule
    Subset & \multicolumn{2}{c}{15} & \multicolumn{2}{c}{13} & \multicolumn{2}{c}{8} & 4 & 8 & 6 \\
    All tasks & \multicolumn{2}{c}{48} & \multicolumn{2}{c}{37} & \multicolumn{2}{c}{21} & 14 & 21 & 19 \\
    \bottomrule
  \end{tabular}
\end{table}

\section{Verifiers and Scoring}
\label{app:scoring}

\paragraph{Verifiers.}
Every task version has its own verifier, a program that reads the final state of the world rather than judging screenshots.
For computer use, it reads the application's records and the file system, \eg the messages sent, the files created and their contents, or the cells opened on a board.
For robot use, it reads object poses, contacts, and joint states in the simulator, \eg which tier each item rests on or whether a drawer is open.
A run succeeds only if the goal is reached and every check of the rule version given in the context passes.

\paragraph{Process scores.}
The process score $p_j$ gives partial credit to failed runs:
\begin{itemize}[nosep,leftmargin=*]
  \item \textbf{Easy computer-use tasks} apply the rule to several objects; $p_j$ is the fraction of objects handled correctly.
  \item \textbf{Medium and hard computer-use tasks} embed the rule as one module of a longer workflow; $p_j=0.4$ if the rule-bearing step is verified on the saved state but the workflow is incomplete, and $p_j=0$ otherwise.
  \item \textbf{Robot tasks} are decomposed into weighted milestones (Fig.~\ref{fig:framework}); a failed run receives the weighted fraction of the milestones it achieves, capped at $0.9$ so that it stays below a success.
\end{itemize}

\section{Process Scores and Per-Task Results}
\label{app:detail}

\paragraph{Process scores.}
\cref{tab:process} reports the process scores of the main experiment.
As with success, the strongest agents earn most of the credit in computer use but only about a third in robot use.

\paragraph{Per-task results.}
At the level of individual tasks, the results are consistent across models and modalities.
In both domains, every task that one of the other four models solves is also solved by Claude Opus 5.5 or GPT-6 Astra with the same rule modality, so the tasks form a common ordering of difficulty rather than model-specific outcomes.
With text and with video, Claude Opus 5.5 reaches the same outcome on 91\% of the computer-use tasks and GPT-6 Astra on 84\%, so their small text--video gap (Sec.~\ref{sec:analysis}) does not arise from large task-level differences that cancel out.

\begin{table}[!htb]
  \centering
  \scriptsize
  \setlength{\tabcolsep}{4pt}
  \caption{\textbf{Process scores} (rule C, text and video pooled). \emph{Mean}: mean process score over all runs (\%); \emph{Partial}: share of failed runs with a positive process score (\%).}
  \label{tab:process}
  \begin{tabular}{@{}l cc cc@{}}
    \toprule
    & \multicolumn{2}{c}{Computer use} & \multicolumn{2}{c}{Robot use} \\
    \cmidrule(lr){2-3}\cmidrule(l){4-5}
    Model & Mean & Partial & Mean & Partial \\
    \midrule
    Claude Opus 5.5 & 84.5 & 27.5 & 32.3 & 16.4 \\
    GPT-6 Astra & 79.8 & 26.9 & 31.1 & 14.9 \\
    Qwen3.8-Max & 42.1 & 26.5 & 12.3 & 13.3 \\
    DeepSeek-V4.1-Flash & 29.7 & 20.2 & 13.1 & 12.3 \\
    GLM-5.3-Flash & 38.7 & 20.7 & 5.3 & 9.9 \\
    Kimi K3 & 36.6 & 23.5 & 3.2 & 8.4 \\
    \bottomrule
  \end{tabular}
\end{table}

\section{Successes Without a Rule and Rule Versions}
\label{app:norule}

\paragraph{Successes without a rule.}
Without the rule, the three ablation models succeed on only a few computer-use runs and on no robot run (Tab.~\ref{tab:abl_norule}).
\cref{tab:norule_cases} explains these runs: in each, the agent picks a plausible behavior from what the interface shows, and this behavior happens to coincide with the rule.
None of them reflects knowledge of the rule.

\begin{table}[!htb]
  \centering
  \scriptsize
  \setlength{\tabcolsep}{3pt}
  \caption{\textbf{Why the few runs without a rule succeed.}}
  \label{tab:norule_cases}
  \begin{tabular}{@{}>{\raggedright\arraybackslash}p{0.25\linewidth}>{\raggedright\arraybackslash}p{0.7\linewidth}@{}}
    \toprule
    Task & What happened \\
    \midrule
    Merge conflicting records (3 models) & A file of field timestamps is in the workspace; the models keep the newer value of each field, which coincides with the rule. \\
    \addlinespace
    Deliver files (1 model) & The output folder offers several groupings; the model picks the one that best matches the goal, which coincides with the rule. \\
    \addlinespace
    Add conversation shortcuts (1 model) & The interface shows unread counts; the model picks the conversation with the most unread messages, which coincides with the rule. \\
    \addlinespace
    Open a safe Minesweeper cell (1 model) & The model takes the first safe cell in reading order, which coincides with the rule on all three boards. \\
    \bottomrule
  \end{tabular}
\end{table}

\paragraph{Rule versions.}
No version is uniformly easier: version C is the easiest for GPT-6 Astra in computer use but the hardest for DeepSeek-V4.1-Flash and GLM-5.3-Flash (Tab.~\ref{tab:abl_norule}).
Which version an agent follows thus depends on how the rule meets the agent.
Two examples illustrate the properties discussed in Sec.~\ref{sec:ablation}.
In a 2048 task, the rules to maximize the merged tiles or the score agree with common objectives of the game, whereas the rule to keep the largest tile in a corner departs from them; models succeed far more often on the former.
In a task that labels problem difficulty, rules based on the number of examples shown on the screen are followed more often than a rule based on the parity of the problem ID, which the interface does not suggest.

\section{Termination in Robot Use}
\label{app:termination}

A robot episode ends when the agent presses \emph{Done}, gives up, or reaches the step limit.
\cref{tab:termination} shows how failed robot runs end.
GPT-6 Astra and Qwen3.8-Max usually give up when they cannot complete a task, whereas Claude Opus 5.5 and GLM-5.3-Flash more often declare completion although the task is unfinished, \ie they misjudge the final state in the camera views; the step limit rarely ends a run.
Together with the rule questions (Fig.~\ref{fig:qa}), this shows that GPT-6 Astra usually knows the rule and, when it fails, recognizes that its execution falls short, which locates the gap in execution rather than in understanding.

\begin{table}[!htb]
  \centering
  \scriptsize
  \setlength{\tabcolsep}{4pt}
  \caption{\textbf{How failed robot runs end} (share of failed runs, \%).}
  \label{tab:termination}
  \begin{tabular}{@{}lccc@{}}
    \toprule
    Model & Gives up & Declares done & Step limit \\
    \midrule
    GPT-6 Astra & 76 & 24 & 0 \\
    Qwen3.8-Max & 73 & 26 & 1 \\
    DeepSeek-V4.1-Flash & 55 & 45 & 0 \\
    Kimi K3 & 50 & 45 & 4 \\
    Claude Opus 5.5 & 44 & 56 & 0 \\
    GLM-5.3-Flash & 41 & 56 & 3 \\
    \bottomrule
  \end{tabular}
\end{table}

\section{Case Study}
\label{app:case}

\begin{figure}[!htb]
  \centering
  \includegraphics[width=\linewidth]{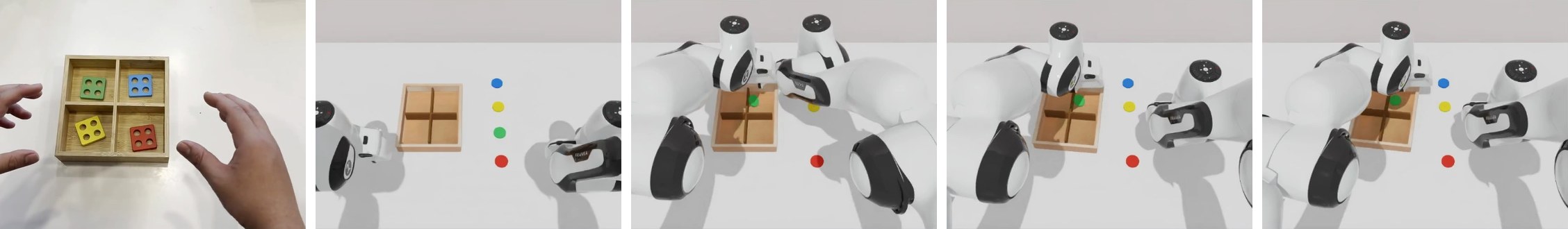}
  \caption{\textbf{Identified but not executed.} Left: end of the real-world demonstration. Right: head-camera views of GPT-6 Astra's run in simulation, from the initial scene to the moment it gives up.}
  \label{fig:case}
\end{figure}

\cref{fig:case} shows a run of GPT-6 Astra on a robot task in which four blocks must be loaded into a 2$\times$2 box in a demonstrated color order.
Before acting, the agent answers the rule question correctly, and it starts with the first block of the demonstrated order.
It lowers the block into the box, but the block does not settle into its cell; the agent repeatedly tries to re-grasp it, then gives up, reporting that it cannot correct the block's position.
The agent knew what to do and failed in how to do it, the pattern behind the gap between understanding and execution.

\section{Confidence Intervals}
\label{app:ci}

\cref{tab:ci} gives the 95\% bootstrap confidence intervals over tasks for the comparisons of Sec.~\ref{sec:analysis}.
The intervals of the frontier agents' text--video gaps include zero, whereas the gap of the other four models excludes it; from \vR to \vI, video-based success drops, and it drops by more than text-based success.

\begin{table}[!htb]
  \centering
  \scriptsize
  \setlength{\tabcolsep}{4pt}
  \caption{\textbf{Bootstrap 95\% confidence intervals} (percentage points; 10{,}000 resamples of tasks, rule C). Gaps are text minus video; level changes are the later level minus the earlier one on the same tasks. Intervals are not adjusted for multiple comparisons.}
  \label{tab:ci}
  \begin{tabular}{@{}l r c@{}}
    \toprule
    Comparison & Estimate & 95\% CI \\
    \midrule
    \multicolumn{3}{@{}l}{\emph{Text--video gap, computer use}} \\
    \quad Claude Opus 5.5 & 0.0 & [$-$5.5, 5.5] \\
    \quad GPT-6 Astra & $-$1.8 & [$-$9.1, 5.5] \\
    \quad Other four models, pooled & 6.1 & [1.8, 10.7] \\
    \addlinespace
    \multicolumn{3}{@{}l}{\emph{Text--video gap by rule type, all models, both domains}} \\
    \quad Procedure & 9.6 & [2.6, 18.4] \\
    \quad Labeling and matching & 7.9 & [0.8, 15.9] \\
    \quad Selection & 4.5 & [0.3, 9.0] \\
    \quad Ordering & $-$1.2 & [$-$13.1, 9.5] \\
    \quad Arrangement & $-$1.6 & [$-$7.1, 3.2] \\
    \quad Formatting & $-$2.7 & [$-$9.9, 4.1] \\
    \addlinespace
    \multicolumn{3}{@{}l}{\emph{\vR $\rightarrow$ \vI, computer use, all models (13 tasks)}} \\
    \quad Video & $-$29.5 & [$-$43.6, $-$16.7] \\
    \quad Text & $-$1.3 & [$-$19.2, 17.9] \\
    \quad Video minus text & $-$28.2 & [$-$48.7, $-$9.0] \\
    \addlinespace
    \multicolumn{3}{@{}l}{\emph{\vI $\rightarrow$ \vE, computer use, frontier agents (37 tasks)}} \\
    \quad Video & 0.0 & [$-$8.1, 6.8] \\
    \quad Text & $-$2.7 & [$-$9.5, 4.1] \\
    \bottomrule
  \end{tabular}
\end{table}

\end{document}